\documentclass{article}

\usepackage{arxiv}

\usepackage[utf8]{inputenc} 
\usepackage[T1]{fontenc}    
\usepackage{hyperref}       
\usepackage{url}            
\usepackage{booktabs}       
\usepackage{amsfonts}       
\usepackage{nicefrac}       
\usepackage{microtype}      
\usepackage{lipsum}

\usepackage{amsmath, nccmath}
\usepackage[numbers]{natbib}
\usepackage{graphicx}
\usepackage{amsfonts}
\usepackage[dvipsnames]{xcolor}
\usepackage{tabularx}

\newcommand{\pp}[1]{\left( #1 \right)}

\newcommand{\mc}{\mathcal}

\usepackage{color}
\definecolor{darkgreen}{rgb}{0,0.6,0}

\newcommand{\papertitle}{
On the infinite width limit of neural networks with a standard parameterization
}

\usepackage{hyperref}
\hypersetup{
    unicode=false,
    pdftoolbar=true,
    pdfmenubar=true,
    pdffitwindow=false,
    pdfstartview={FitH},
    pdftitle={\papertitle},
    pdfauthor={Jascha Sohl-Dickstein},
    pdfsubject={\papertitle},
    pdfcreator={},
    pdfproducer={},
    pdfkeywords={},
    pdfnewwindow=true,
    colorlinks=True,
    linkcolor=BurntOrange,
    citecolor=RawSienna,
    filecolor=magenta,
    urlcolor=blue
}

\title{\papertitle
}

\newcommand{\email}[1]{\tt\href{mailto:#1@google.com}{#1}}

\author{
  Jascha Sohl-Dickstein,\,\,\,\,\,\, Roman Novak,\,\,\,\,\,\, Samuel S. Schoenholz,\,\,\,\,\,\, Jaehoon Lee \\
  \texttt{\{\email{jaschasd}, \email{romann}, \email{schsam}, \email{jaehlee}\}@google.com}
}

\begin{document}
\maketitle



\begin{abstract}
    There are currently two parameterizations used to derive fixed kernels corresponding to infinite width neural networks, the NTK (Neural Tangent Kernel) parameterization and the naive standard parameterization. However, the     extrapolation of both of these parameterizations to infinite width is problematic. The standard parameterization leads to a divergent neural tangent kernel while the NTK parameterization fails to capture crucial aspects of finite width networks such as: the dependence of training dynamics on relative layer widths, the relative training dynamics of weights and biases, and overall learning rate scale. 
    Here we propose an improved extrapolation of the standard parameterization that preserves all of these properties as width is taken to infinity and yields a well-defined neural tangent kernel. 
    We show experimentally that the resulting kernels typically achieve similar accuracy to those resulting from an NTK parameterization, but with better correspondence to the parameterization of typical finite width networks. 
    Additionally, with careful tuning of width parameters, the improved standard parameterization kernels can outperform those stemming from an NTK parameterization.
    We release code implementing this improved standard parameterization as part of the Neural Tangents library \cite{novak2020neural} at  \url{https://github.com/google/neural-tangents}.
\end{abstract}

\section{Introduction}

\begin{figure}
     \centering
     \includegraphics[width=0.85\linewidth]{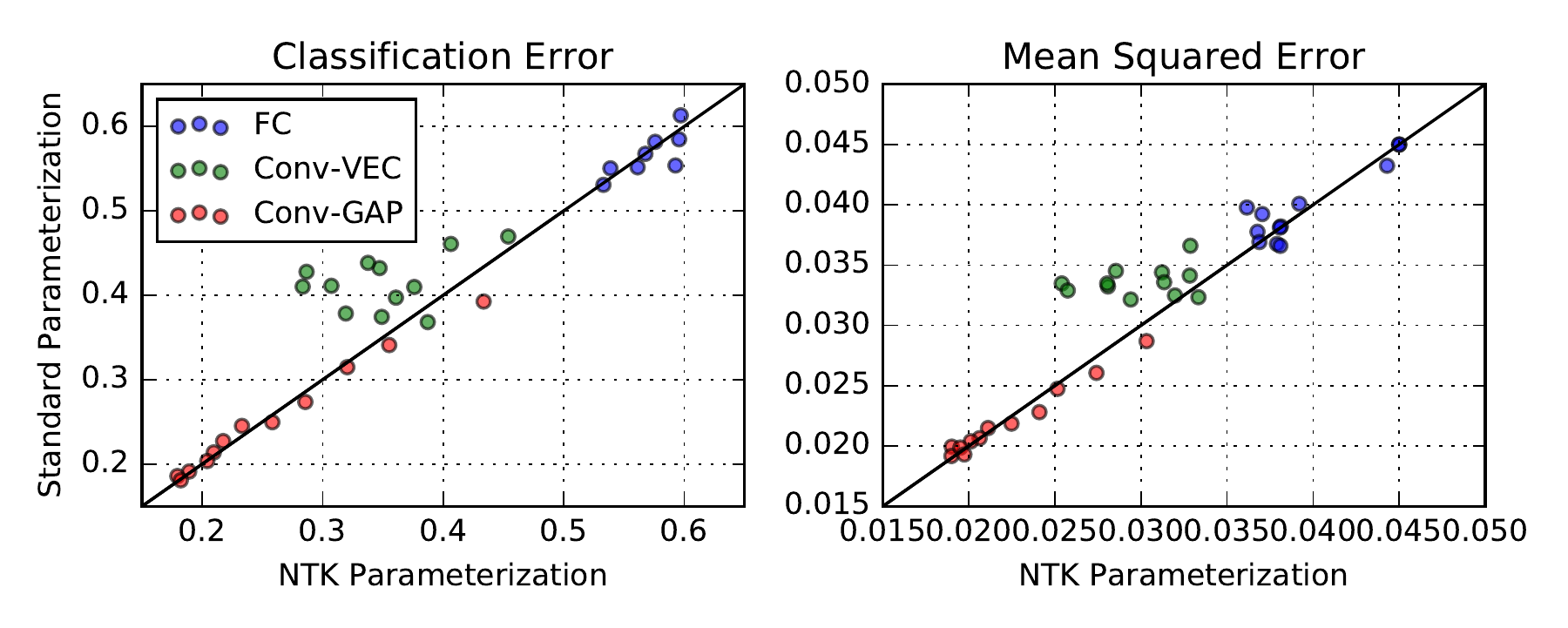}
     \caption{SGD trained finite width neural networks perform similarly when using the standard parameterization or the NTK parameterization. For all experiments, the network was trained with an MSE loss on the full CIFAR-10 dataset (45k/5k/10k split). 
     Each point in \textbf{FC} corresponds to varying width $\{2^k| k=0,...,12\}$, and each point in \textbf{Conv-VEC} and \textbf{Conv-GAP} corresponds to varying number of channels \{8, 11, 16, 23, 32, 45, 64, 90, 128, 181, 256, 362, 512\}.
     All networks are ReLU networks with $\sigma_w^2 = 2.0, \sigma_b^2=0.0$. They were trained with vanilla SGD without L2 regularization or data augmentation. Constant learning rate was grid searched over 20 log spaced values within [0.01, 100]. For standard parameterization learning rate is divided by $\max(N^l)$.
     \textbf{FC} networks were trained with batch size 1024 for 3,000 epochs whereas \textbf{Conv} networks were trained with batch size 256 for 10,000 epochs. }
     \label{fig:scatter_nn}
 \end{figure}

Infinite width Bayesian \cite{neal,lee2018deep, matthews2018,matthews2018b_arxiv, novak2018bayesian,garriga-alonso2018deep,NIPS2019_8809,yang2019scaling,yang2019wide,de2019random} and gradient descent trained \cite{jacot2018neural,lee2019wide,chizat2019lazy,yang2019scaling,jacot2019freeze,dyer2019asymptotics,bietti2019inductive,arora2019on,arora2019harnessing,schwartz2019information} neural networks are an area of active and extremely promising work. 
There are currently two parameterizations used to derive fixed kernels corresponding to infinite width neural networks\footnote{Another line of work applies a different scaling, and derives non-fixed infinite width kernels \cite{mei2018mean,mei2019mean,chizat2018global,Nguyen2019MeanFL}.}: the NTK parameterization \cite[\S 2]{jacot2018neural}; and the naive standard parameterization \cite[\S 2.1]{park2019effect}; \cite{glorot2010understanding, he2016deep}. 
However, the extrapolations of both of these parameterizations to infinite width fail to capture crucial aspects of finite width networks:
\begin{itemize}
    \item In finite width networks, differences in relative layer widths can have a profound effect on training dynamics. Under the NTK parameterization, as layer width goes to infinity, relative layer width has no effect on training dynamics or predictions.
    \item As the naive standard parameterization is extended to large widths, the largest stable learning rate scales like $\frac{1}{\text{width}}$ \cite[Theorem 7]{karakida2018universal}; \cite[\S H]{park2019effect}. A learning rate that goes to zero as width goes to infinity poses a variety of practical and theoretical challenges, including a neural tangent kernel with entries that diverge to infinity.
    \item At finite width, convolutional networks with an NTK parameterization have been reported to generalize more poorly than those with a standard parameterization \cite[\S I]{park2019effect} (though we do not consistently reproduce this relationship 
    in our own experiments, see Figure \ref{fig:scatter_nn}).
    \item For neither NTK nor naive standard parameterizations do infinite width learning rates agree closely with those typically used to train finite width standard parameterization networks.
    \item The relative learning dynamics of bias and weight parameters are different in the NTK parameterization than they are for a standard parameterization finite-width network.
\end{itemize}

In this note we propose an improved extrapolation of the standard parameterization to infinite width that resolves these inconsistencies while simultaneously leading to a well-defined neural tangent kernel. Namely, in this parameterization the resulting infinite width network
maintains a learning rate scale that agrees with that used to train the original network, preserves the impact of relative layer widths on training dynamics for finite width networks, and similarly preserves the relative training dynamics of weights and biases.

\begin{table}
\begin{center}
\begin{tabularx}{\textwidth}{|X|X|X|X|} \hline
Parameterization & Standard (naive) & NTK & \textbf{Standard (improved)} \\\hline
Layer equation,\hfill$x^{l+1} = $ & 
    $W^lx^l + b^l$ & 
    $\frac{\sigma_w}{\sqrt{s N^l}}W^lx^l + \sigma_b b^l$  & 
    $\frac{1}{\sqrt{s}}W^lx^l + b$  \\\hline
Weight shape, \hfill $W^l \in $& 
\multicolumn{3}{l|}{$\mc R^{sN^{l+1} \times sN^l}$} \\\hline
$W$ initialization,\hfill$W_{ij}^l \sim$ & 
    $\mc N\left(0, \frac{\sigma_w^2}{sN^l}\right)$ & 
    $\mc N(0, 1)$ & 
    $\mc N\left(0, \frac{\sigma_w^2}{N^l}\right)$ \\\hline
$b$ initialization,\hfill$b_i^l \sim$ & 
    $\mc N\left(0, \sigma_b^2\right)$ &
    $\mc N(0, 1)$ & 
    $\mc N\left(0, \sigma_b^2\right)$ \\\hline
NNGP,\hfill $s \rightarrow \infty, \,K^{l+1} =$ & 
\multicolumn{3}{l|}{
    $\sigma_w^2 K^l + \sigma_b^2$ }
    \\\hline
NTK,\hfill $s \rightarrow \infty, \,\Theta^{l+1} = $ & 
    diverges &
    $\sigma_w^2 K^l + \sigma_b^2 + \sigma_w^2 \Theta^l $ & 
    $N^l K^l + 1 + \sigma_w^2 \Theta^l $ \\\hline
\end{tabularx} 
\end{center}
\caption{
Equations describing a \textbf{fully connected layer} for each parameterization, both for a finite width network and for the corresponding infinite width NNGP and NT kernels. 
Here $N^l$ is the baseline (finite network) width of layer $l$, and $s$ is a width-scaling factor that is taken to $\infty$ for infinite width networks. 
}
\label{tab FC parameterizations}
\end{table}

\begin{table}
\begin{center}
\begin{tabular}{|l|l|l|l|} \hline
Parameterization & Standard (naive) & NTK & \textbf{Standard (improved)} \\\hline
Layer equation,\hfill $x^{l+1}_{i,p} =$ & 
    $W^l_{i,j,m} x^l_{j,p+m} + b_i^l$
    & 
    $\frac{\sigma_w}{\sqrt{s N^l M}} W^l_{i,j,m} x^l_{j,p+m} + \sigma_b b_i^l$    
     & 
    $\frac{1}{\sqrt{s}} W^l_{i,j,m} x^l_{j,p+m} + b_i^l$
  \\\hline
Weight shape,\hfill $W^l\in$ & 
\multicolumn{3}{l|}{$\mc R^{sN^{l+1} \times sN^l \times M}$} \\\hline
$W$ initialization,\hfill $W^l_{i j m}\sim$ & 
    $\mc N\left(0, \frac{\sigma_w^2}{s N^l M}\right)$ & 
    $\mc N(0, 1)$ & 
    $\mc N\left(0, \frac{\sigma_w^2}{N^l M}\right)$ \\\hline
$b$ initialization,\hfill $b_i^l \sim$ & 
    $\mc N\left(0, \sigma_b^2\right)$ &
    $\mc N(0, 1)$ & 
    $\mc N\left(0, \sigma_b^2\right)$ \\\hline
NNGP,\hfill $s \rightarrow \infty, \,K^{l+1} = $ & \multicolumn{3}{l|}{
    $\sigma_w^2 \mathcal A\pp{K^l} + \sigma_b^2$}
    \\\hline
NTK,\hfill $s \rightarrow \infty, \,\Theta^{l+1} = $ & 
    diverges &
    $\sigma_w^2 \mathcal A\pp{K^l} + \sigma_b^2 + \sigma_w^2 \mathcal A\pp{\Theta^l} $ & 
    $N^l M \mathcal A\pp{K^l} + 1 + \sigma_w^2 \mathcal A\pp{\Theta^l} $ \\\hline
\end{tabular} 
\end{center}
\caption{
Equations describing a \textbf{convolutional layer} for each parameterization, both for a finite width network and for the corresponding infinite width NNGP and NT kernels. We use Einstein notation for summation -- indices that appear only in a single term are implicitly summed over. $M$ is the number of spatial positions in the convolution kernel, $m$ indexes over spatial locations within the kernel, $p+m$ corresponds to input spatial location $p$ offset by $m$, $N^l$ is the baseline (finite network) channel count of layer $l$, $\mathcal A\pp{\cdot}$ is the diagonal averaging operator defined in \citet{xiao18a} and \citet[\S 2.2.1]{novak2018bayesian}, and $s$ is a width-scaling factor that is taken to $\infty$ for infinite channel count networks. 
}
\label{tab CNN parameterizations}
\end{table}

 \begin{figure}
     \centering
     \includegraphics[width=0.85\linewidth]{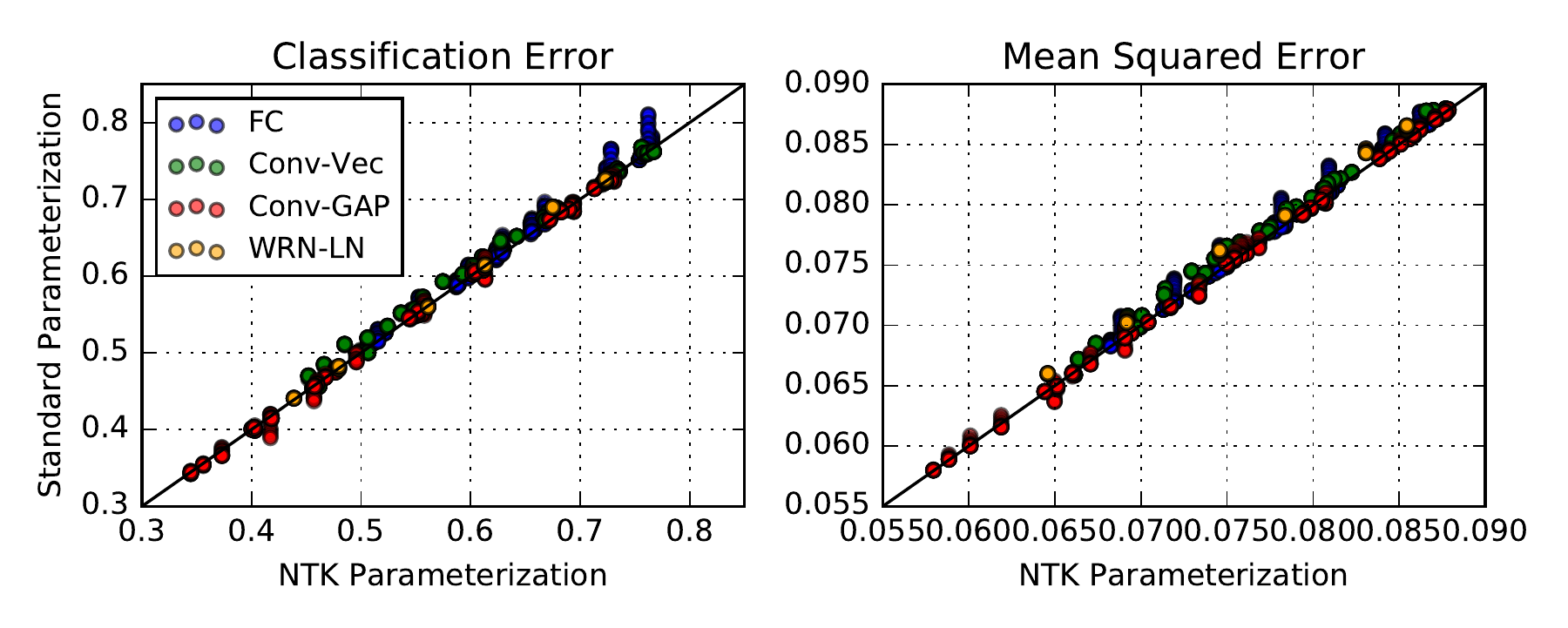}
     \includegraphics[width=0.85\linewidth]{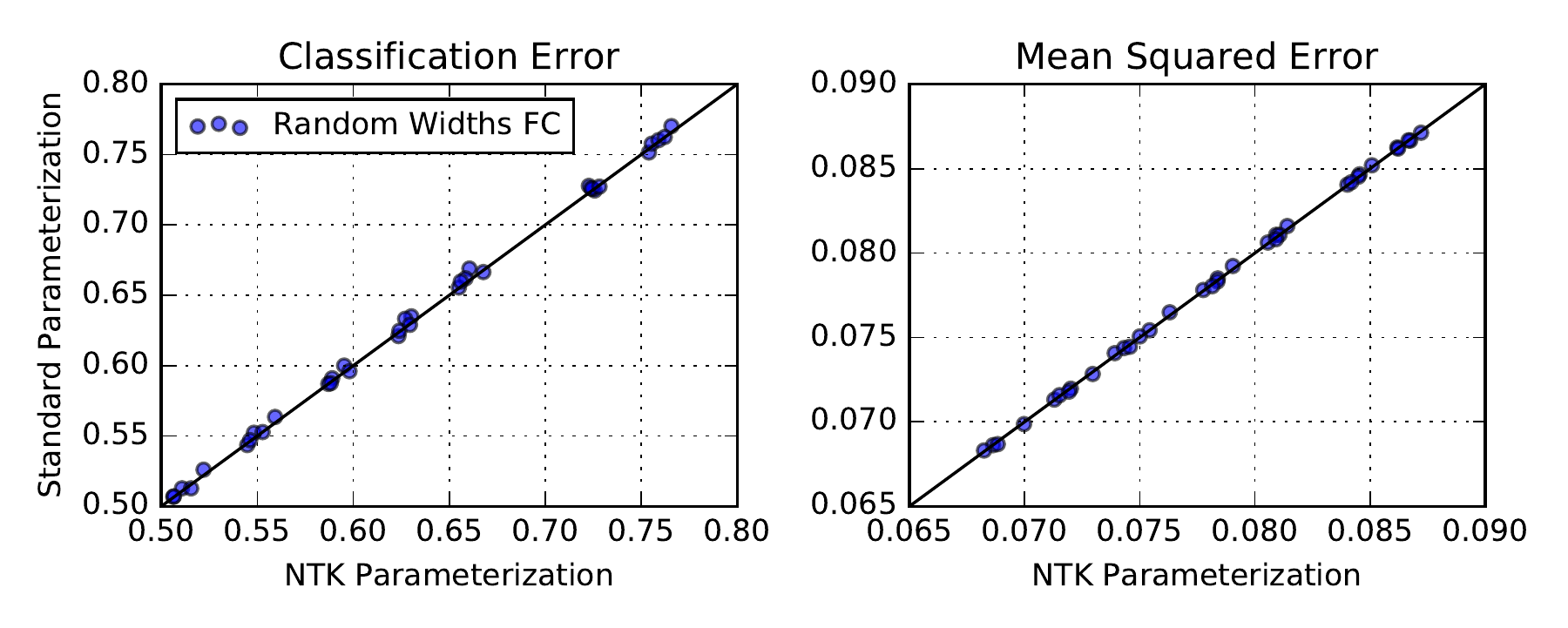}
     
     \caption{Infinite width networks with various architectures 
     achieve similar error 
     when using the improved standard parameterization or the NTK parameterization, while the improved standard parameterization better matches properties of typical finite width networks. Each point compares the neural tangent kernel prediction error for the same architecture on CIFAR-10, but using NTK (x-axis) or improved standard (y-axis) parameterization. {\em (Upper)} Each point corresponds to varying training set size ($\{80, 160, 400, 800, 2000, 4000, 8000\}$), depth ($\{1, 2, 4, 8, 16\}$ for FC / Conv, fixed number of block of 4 for WRN) and widths ($\{2^k| k=0,...,13\}$ for FC / Conv and widening factor $\{2^k|k=-4,...2\}\cup \{10, 16, 64, 256\}$ for WRN). 
     \textbf{FC} is fully connected network with constant hidden width and \textbf{Conv-Vec / GAP} correspond to constant channel convolutional neural networks without / with global average pooling. \textbf{WRN-LN} is Wide Residual Network with four residual blocks and Batch Normalization layer replaced with Layer Normalization.  {\em (Lower)} Each layer width of fully connected architecture are randomly sampled from $2^k$ with $k\in\{3,..., 13\}$.} 
     \label{fig:scatter}
 \end{figure}
  
 \begin{figure}
     \centering
     \includegraphics[width=0.85\linewidth]{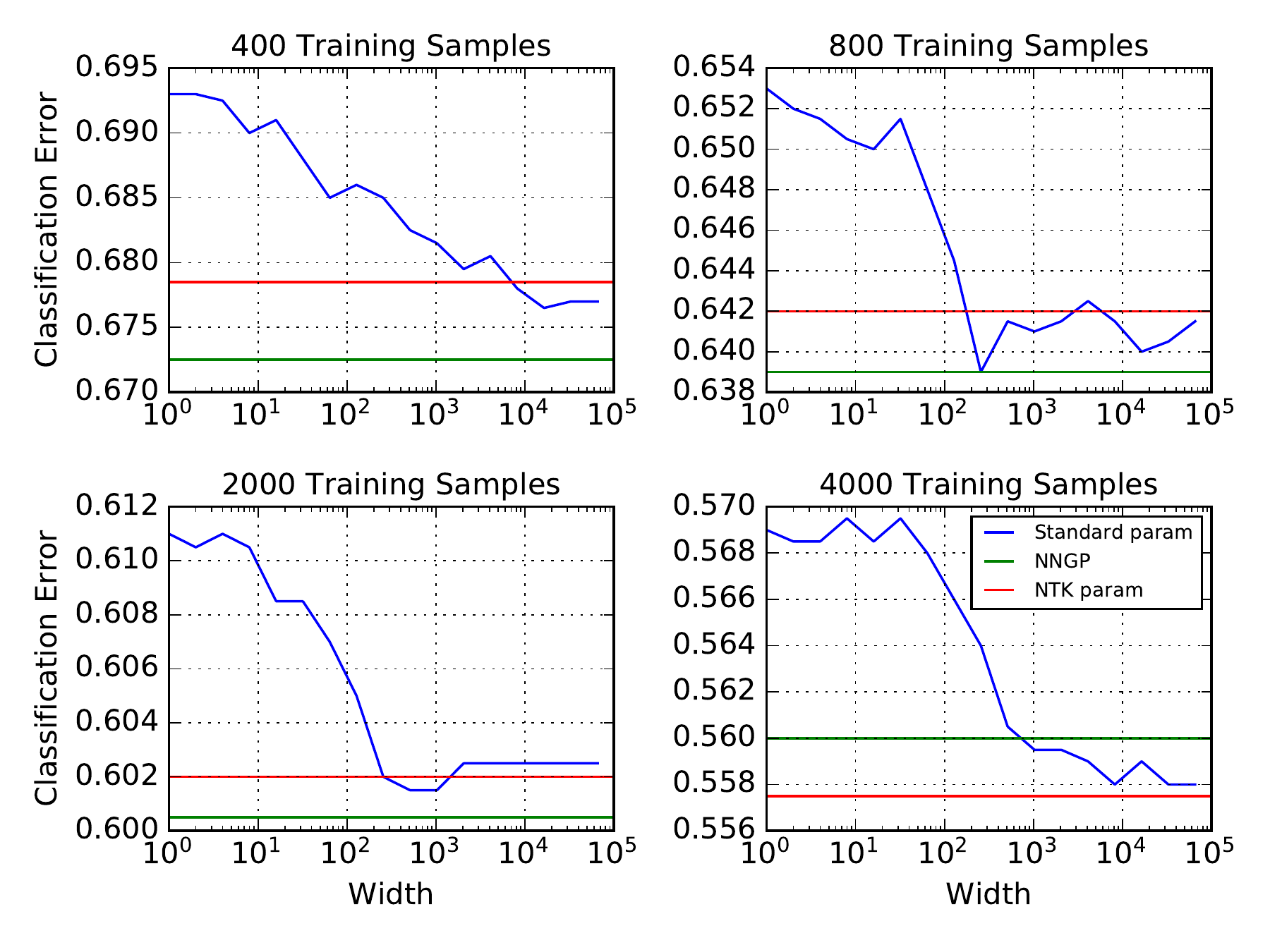}
     \caption{For fully connected networks, the neural tangent kernel prediction for the improved standard parameterization can outperform the NTK parameterization, especially when the layer widths $N^l$ used in the standard parameterization are tuned. Experiments are performed on the CIFAR-10 dataset with networks corresponding to 5 hidden layers. } 
     \label{fig:sweep}
 \end{figure}
 
\section{Improved standard parameterization}

Affine layers in neural networks are typically written as,
\begin{equation}
    z^{l + 1} = W^l y^l + b^l
\end{equation}
where $z^l$ are pre-activations, $y^l = \phi(z^l)$ are activations, $W^l$ are weights, and $b^l$ are biases. To preserve the scale of the pre-activations as the width of the network, $N^l$, is varied one typically initializes the weights as $W^l\sim \mathcal N(0, \sigma_w^2 / N^l)$ and biases as $b^l \sim \mathcal N(0, \sigma_b^2)$. However, as was noted in~\cite{jacot2018neural}, this leads to divergent gradient flow dynamics as $N^l\to\infty$. In~\cite{jacot2018neural}, the authors resolve this situation by using an alternative parameterization where affine layers are written as,
\begin{equation}
    z^{l + 1} = \frac{\sigma}{\sqrt{N^l}} \omega^l y^l + b^l
\end{equation}
where $\omega^l\sim\mathcal N(0, 1)$. This leads to a well-behaved infinite-width limit, but involves a number of inconsistencies relative to standard neural networks.

The core idea here is to write the width of the neural network in each layer in terms of an auxiliary parameter, $s$, $n^l = sN^l$. We then write an affine layer as,
\begin{equation}
    z^{l + 1} = \frac1{\sqrt s}W^l y^l + b^l
\end{equation}
The infinite width limit can be taken by letting $s\to\infty$. The parameter variances $\sigma^2_w, \sigma^2_b$ and original layer widths $N^l$ instead appear in the variance of the initializer (as is typically done for finite width networks). A complete set of equations describing an affine layer, and corresponding infinite width kernels, for this parameterization are given in Tables \ref{tab FC parameterizations} and \ref{tab CNN parameterizations}, for fully connected and convolutional architectures respectively. 

A formal proof of convergence of the improved standard parameteriation to the specified kernels is beyond the scope of this short note. However, we observe that the proof technique in \citet[\S Apps. F,  G]{lee2019wide} applies with minimal modification. Additionally, Monte Carlo validation of the correctness of the introduced kernels is performed as part of the Neural Tangents \cite{novak2020neural} unit test suite.

\section{Experiments}

In this section, we study empirical properties of infinite and finite width networks stemming from both the NTK and improved standard parameterization.
All of the experiments in this section were done using Neural Tangents library~\cite{novak2020neural}. Here we focus on kernels corresponding to ReLU networks with $\sigma_w^2 = 2.0, \sigma_b^2=0.1$.

In Figure \ref{fig:scatter} we compare the predictions of kernels for pairs of identical networks, but using the improved standard or NTK parameterization. We find that the performance of the kernels resulting from the two parameterizations are extremely similar, while the training dynamics of the improved standard parameterization network are expected to better match those of typical finite width networks. In Figure \ref{fig:sweep} we show that if the width parameter $N^l$ is carefully tuned, then the neural tangent kernel for a fully connected network using the improved standard parameterization can outperform the kernel for an NTK parameterized network. In Figure \ref{fig:scatter_nn}, we show that random finite width networks using the standard and NTK parameterization perform similarly.

 \section{Discussion}

The analytic forms for the various kernels inspire some additional interesting observations:
\begin{itemize}
     \item For the NTK parameterization, the kernel resulting from a Bayesian neural network and from gradient descent training of the readout layer of an infinite width network are the same. For the both the naive and improved standard parameterization however, the two differ.
     \item For neural networks with a standard parameterization, the magnitude of the contribution of the bias to the neural tangent kernel (and thus to learning dynamics) remains constant with increasing width. However, the contribution of the weights to the learning dynamics grows like like $N^l$. We should thus expect that as networks become wide, the role played by the bias in training becomes less important.
\end{itemize}

In this note, we introduced an improved extrapolation of finite width networks to infinite width that better matches the parameterization and learning dynamics of typical finite width networks. It is our hope that this will enable theory and experiments with infinite width networks to better explain the behavior of practical finite width networks.

\bibliography{references}
\bibliographystyle{apalike}

\end{document}